\documentclass[letterpaper]{article} 
\usepackage{aaai2026}  
\nocopyright
\usepackage{times}  
\usepackage{helvet}  
\usepackage{courier}  
\usepackage[hyphens]{url}  
\usepackage{graphicx} 
\usepackage{natbib}  
\usepackage{caption} 
\usepackage{algorithm}
\usepackage{algorithmic}
\usepackage{placeins} 
\usepackage{newfloat}
\usepackage{listings}
\DeclareCaptionStyle{ruled}{labelfont=normalfont,labelsep=colon,strut=off} 
\floatstyle{ruled}
\newfloat{listing}{tb}{lst}{}
\floatname{listing}{Listing}
\title{Text-based Aerial-Ground Person Retrieval}
\author{
    Xinyu Zhou \textsuperscript{\rm 1},
    Yu Wu \textsuperscript{\rm 1},
    Jiayao Ma \textsuperscript{\rm 2},
    Wenhao Wang \textsuperscript{\rm 2},
    Min Cao \thanks{Corresponding Author} \textsuperscript{\rm 1},
    Mang Ye \textsuperscript{\rm 3}
}
\affiliations{
    \textsuperscript{\rm 1}School of Computer Science and Technology, Soochow University\\
    \textsuperscript{\rm 2}AgiBot,
    \textsuperscript{\rm 3}School of Computer Science, Wuhan University\\
    xyzhou2023@stu.suda.edu.cn, mcao@suda.edu.cn

}

\usepackage{bibentry}

\usepackage{xcolor}
\usepackage{colortbl}
\usepackage{multirow}
\usepackage{tcolorbox}
\definecolor{lightgray}{gray}{.9}
\usepackage{xspace}
\usepackage{pifont}
\usepackage{amssymb}
\usepackage{amsmath} 
\usepackage{mathtools}

\begin{document}

\maketitle

\begin{abstract}

This work introduces Text-based Aerial-Ground Person Retrieval (TAG-PR), which aims to retrieve person images from heterogeneous aerial and ground views with textual descriptions. 
Unlike traditional Text-based Person Retrieval (T-PR), which focuses solely on ground-view images, TAG-PR introduces greater practical significance and presents unique challenges due to the large viewpoint discrepancy across images.  
To support this task, we contribute: (1) TAG-PEDES dataset, constructed from public benchmarks with automatically generated textual descriptions, enhanced by a diversified text generation paradigm to ensure robustness under view heterogeneity; and (2) TAG-CLIP, a novel retrieval framework that addresses view heterogeneity through a hierarchically-routed mixture of experts module to learn view-specific and view-agnostic features and a viewpoint decoupling strategy to decouple view-specific features for better cross-modal alignment.
We evaluate the effectiveness of TAG-CLIP on both the proposed TAG-PEDES dataset and existing T-PR benchmarks. The dataset and code are available at https://github.com/Flame-Chasers/TAG-PR. 

\end{abstract}


\section{Introduction}
Text-based Person Retrieval (T-PR)~\cite{li2017person} is a specialized vision-language learning task that identifies person images using natural language descriptions. 
It can be viewed as an extension of the traditional person re-identification (Re-ID)~\cite{wang2022nformer, liu2024learning}, which matches person images across different cameras, by replacing the query image with text.
T-PR has gained increasing academic attention in recent years~\cite{bai2023text, song2024diverse, tan2024harnessing, bai2025chat} due to its potential in applications such as suspect tracking and surveillance. 
Current research focuses on achieving effective cross-modal alignment between person images and texts, and has made substantial breakthroughs in performance~\cite{qin2024noisy, jiang2025modeling, cao2025multilingual}.

\begin{figure}[t]
  \centering
  \includegraphics[width=0.89\linewidth]{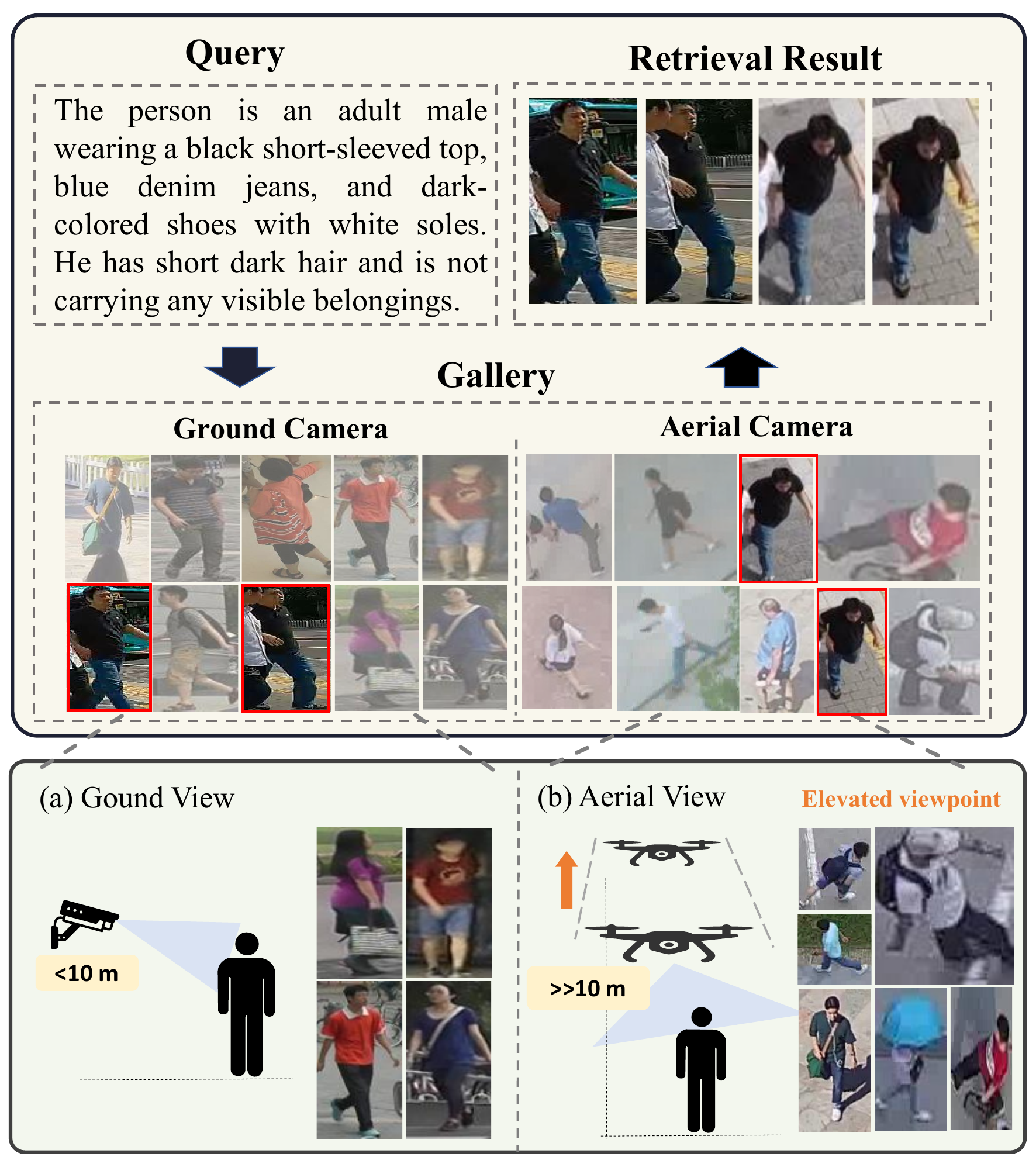}
  \caption{Illustration of TAG-PR. 
  It aims to retrieve a target individual from an image gallery containing heterogeneous-view images, given a query text. The gallery includes ground-view images, typically captured by CCTV cameras at low altitudes (below 10 meters), and aerial-view images taken by UAVs from significantly higher altitudes (\(\gg 10m\)). }
  \label{fig: task}
\end{figure}

Nevertheless, these works focus mainly on scenarios involving only ground-view camera networks, assuming that all person images are captured from low-altitude, ground-view cameras. This assumption does not fully reflect real-world conditions, where advances in airborne platforms and imaging technologies have enabled widespread deployment of aerial cameras alongside traditional ground systems. High-altitude aerial-view cameras offer broader coverage and complementary perspectives to ground-based cameras, and the synergy between these two viewpoints becomes increasingly important in real-world retrieval scenarios.
In response, the Re-ID community has explored in mixed ground-aerial settings~\cite{zhang2023ground, nguyen2024ag, wang2025secap}, such efforts remain absent in T-PR task.
For this, we focus on a more practical setting in this paper, Text-based Aerial-Ground Person Retrieval (TAG-PR), which involves retrieving person images captured from heterogeneous aerial and ground viewpoints based on textual descriptions, as shown in Figure~\ref{fig: task}.

Existing T-PR datasets primarily consist of ground-view person images, lacking datasets specifically tailored for TAG-PR. To address this gap, we firstly introduce TAG-PEDES, a dataset explicitly designed for TAG-PR. It is economically constructed by collecting person images from ground and aerial views based on public Re-ID datasets~\cite{zheng2015person, zhang2020person, li2021uav, nguyen2024ag, zhang2023ground} with automatically generated textual descriptions.
A central challenge lies in generating high-quality descriptions for these images that exhibit significant heterogeneity due to drastic view-angle differences. For this, we propose a Diversified Text Generation (DTG) paradigm powered by multimodal large language models. DTG integrates multiple distinct text generation strategies to ensure robustness and adaptability of high-quality textual descriptions across view-angle images.

In addition, we propose a novel method tailored to this task. 
Traditional T-PR methods focus on cross-modal alignment between image and text. While in more complex TAG-PR, it is equally important to account for view heterogeneity inherent in the image modality. Ground-view images, captured at eye level, present frontal or side views of a person, whereas aerial-view images, taken from high altitudes, offer a top-down perspective, leading to distinct visual characteristics.
For this, we propose TAG-CLIP. It explicitly incorporates view heterogeneity during image encoding and cross-modal alignment.
1) For image encoding, the images from different viewpoints exhibit distinct visual characteristics while also sharing view-agnostic features. A natural way is to extract view-specific patterns separately while jointly learning view-agnostic features. 
For this, we design a Hierarchically-Routed Mixture of Experts (HR-MoE) module and integrate it into the image encoder. It comprises a two-tiered routing mechanism, along with two specialized expert groups, all of which are designed to process view-specific and view-agnostic features separately.
2) For cross-modal alignment learning, unlike images, textual descriptions lack view-specific information\footnote{TAG-PR retrieves target individuals based on textual descriptions that typically focus on appearance-related attributes and do not include viewpoint cues in real-world inference scenarios.}, making it crucial to decouple view-specific features from the image feature before alignment. We propose a viewpoint decoupling strategy that eliminates viewpoint-specific cues from image features. The resulting view-agnostic visual features are then aligned with textual features for better performance.

In summary, our contributions are as follows.
1) We introduce text-based aerial-ground person retrieval, a cross-modal and cross-platform task with broader application potential than traditional text-based person retrieval.
2) We construct a dataset, TAG-PEDES, for this task, supported by the proposed diversified text generation paradigm to collect high-quality textual descriptions from varied viewpoints. 
3) We propose TAG-CLIP, a novel method which incorporates a hierarchically-routed mixture of experts in the image encoder and a viewpoint decoupling strategy to better learn and align view-specific and view-invariant features.

\section{Related Work}
\subsection{Text-based Person Retrieval}
T-PR has emerged as an active research area in recent years, marked by notable methodological advances. Early approaches~\cite{li2017person, wang2017adversarial, chen2021cross, li2017identity} employ independently trained unimodal encoders (e.g., VGG~\cite{simonyan2014very} for images and LSTM~\cite{graves2012long} for text) to extract global features for cross-modal alignment, but struggle to capture fine-grained semantic correspondences.
Subsequent works~\cite{jing2020pose, fujii2023bilma, wang2020vitaa} address this limitation by incorporating explicit alignment mechanisms, often relying on external tools to align fine-grained entities.
More recently, vision-language pretraining models~\cite{radford2021learning, li2021align} have attracted significant attention due to their strong cross-modal representation capabilities. Leveraging these models, a series of studies~\cite{jiang2023cross, cao2024empirical, yan2024prototypical, yang2023towards} have achieved remarkable performance gains. 
For example, Cao \textit{et al.}~\cite{cao2024empirical} conducted comprehensive evaluations of CLIP's fine-grained alignment capacity. 
However, these works focus mainly on text-based retrieval for ground-view images, limiting their real-world applicability.
In contrast, this work introduces TAG-PR.
A concurrent work AEA-FIRM~\cite{wang2025aea} also studies this task but differs:
1) Task Definition. AEA-FIRM defines on strictly paired ground-aerial images per identity, whereas we allow diverse combinations (ground-only, aerial-only, and mixed-view), better reflecting real-world scenarios.
2) Dataset Construction. AEA-FIRM generates textual descriptions solely for ground views and reuses them for aerial views, ignoring viewpoint-specific cues. We propose DTG to enable viewpoint-aware descriptions.
3) Modeling View Heterogeneity. AEA-FIRM employs an AEA loss with a rigid triplet input structure, offering limited flexibility to view differences. We propose TAG-CLIP, which explicitly models view heterogeneity during both image encoding and cross-modal alignment, leading to better performance.

\subsection{Aerial-Ground Person Re-ID}
Aerial-ground person Re-ID has recently emerged as a challenging task. Nguyen \emph{et al.}~\cite{nguyen2023aerial} introduced the first benchmark, AG-ReID, and proposed a two-stream framework combining a transformer-based Re-ID stream with an explainable stream, using attribute supervision to improve feature discriminability.
They later extended the work with the larger AG-ReID.v2 dataset and a more robust three-stream network for improved performance~\cite{nguyen2024ag}.
However, these methods rely on one-hot attribute labels, which may limit generalizability due to their dependence on manual semantic cues.
In parallel, Zhang \emph{et al.}~\cite{zhang2024view} introduced a view-decoupled transformer, yet neglects fine-grained local viewpoint disentanglement crucial for cross-modal alignment.
In contrast, our proposed HR-MoE distinguishes view-specific and view-agnostic local features by assigning them to different expert groups, enabling more robust feature extraction. 

\section{Dataset}

\subsection{Dataset Construction}
To support the TAG-PR task, we construct a benchmark dataset TAG-PEDES, by collecting person images from existing Re-ID datasets and generating corresponding textual descriptions. 
In addition, to ensure test data quality, we manually review and refine the textual annotations of the test set.

\subsubsection{Image Collection.}
To better reflect the complex viewpoint diversity present in real-world scenarios, we curate a collection from multiple Re-ID datasets. Specifically, we obtain ground-view images from ground-only dataset~\cite{zheng2015person}, aerial-view images from aerial-only datasets~\cite{zhang2020person, li2021uav}, and multi-view images from aerial-ground datasets~\cite{nguyen2024ag, zhang2023ground}, with view labels (\emph{i.e.}, aerial or ground) provided in the original datasets. After filtering out low-resolution images ($\leq 1,000$ pixels), we obtain an image gallery containing \(28,206\) images of \(6,840\) unique identities captured from multiple viewpoints and cameras. 
Table~\ref{tab: datasource} presents the detailed composition of the image gallery of TAG-PEDES. 
\begin{figure}[t]
  \centering
  \includegraphics[width=0.98\linewidth, height=0.9\linewidth]{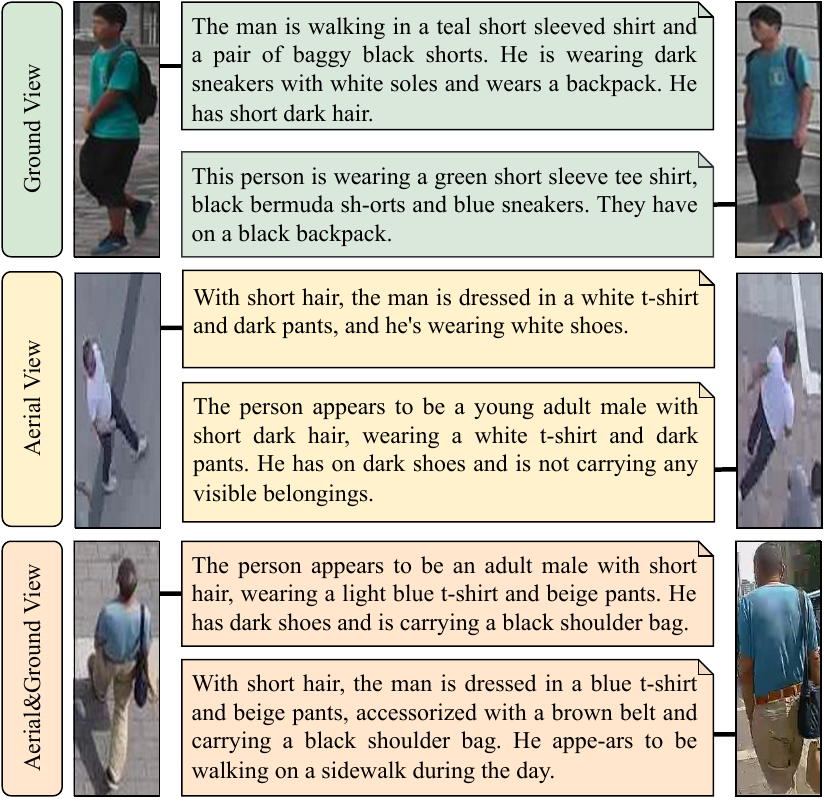}
  \caption{Examples from the proposed TAG-PEDES dataset. More examples are provided in the Appendix.}
  \label{fig: imgDis}
\end{figure}
As shown in Figure~\ref{fig: imgDis}, the diversity in viewpoints introduces variations in image resolution and pedestrian postures, making the dataset more representative of real-world scenarios, while posing challenges for generating high-quality texts.

\begin{table}[t]
    \centering
    \fontsize{9}{11}\selectfont
    \setlength{\tabcolsep}{0.4mm}
    \begin{tabular}{c|ccc}
    \hline

    View     & Image Source   & \#IDs  & \#Images       \\ 
    \hline\hline
    Ground        & Market-1501~\cite{zheng2015person}  & $1,100$ & $3,300$ \\
    \hline
             & PRAI-1581~\cite{zhang2020person}    & $844$ & $3,374$   \\
    \multirow{-2}{*}{Aerial}       
             & UavHuamn~\cite{li2021uav}           & $655$ & $2,476$  \\
    \hline
             & AG-Reid.v2~\cite{nguyen2024ag}      & $1,615$ & $7,750$  \\
    \multirow{-2}{*}{Aerial\&Ground} & G2APS~\cite{zhang2023ground}        
             & $2,626$ & $11,306$  \\
    \hline
    \end{tabular}
    \caption{Composition of the image gallery in TAG-PEDES. All view labels are inherited from the original datasets.}
    \label{tab: datasource}
\end{table}

\subsubsection{Text Generation.}
To address the challenge posed by various image viewpoints, we propose a Diversified Text Generation (DTG) paradigm powered by Multimodal Large Language Models (MLLM)~\cite{li2024llava} to enable automated image annotation. Specifically, DTG comprises three distinct text generation strategies. 

\textbf{Prompt-based generation} is the most straightforward strategy. Specifically, we activate the powerful text-generation capabilities of the MLLM~\cite{wang2024enhancing} with a fixed, empirically designed prompt:

\textit{Don’t mention the background of the people in the image. Please provide a detailed description of this person's age, gender, top (including color and style), bottom (including color and style), hair (including color and style), shoes (including color and style), belongings (including color and style). Finally, combine all the details into a single sentence.}


\textbf{Template-based generation} provides finer control over prompt-based generation by leveraging auxiliary templates to direct the model's attention to key pedestrian attributes. we provide the MLLM with the prompt~\cite{tan2024harnessing}:

\textit{Generate a description about the overall appearance of the person, in a style similar to the template: \{template\}. If some requirements in the template are not visible, you can ignore. Do not imagine any contents not in the image.
} 

The placeholder \textit{\{template\}} is replaced with specific templates presented in the Appendix.

\textbf{Attribute-based generation} provides the MLLM with manually annotated attribute labels (\emph{e.g.}, young, male) from Re-ID datasets, which offer explicit descriptions of pedestrian's age, gender, clothing and so on.
These attributes are integrated into the prompt of the template-based generation to improve the accuracy and detail of textual descriptions:

\textit{Generate a description about the overall appearance of the person,  based on the attribute: [{label}], in a style similar to the template: [{template}]. If some requirements in the template are not visible, you can ignore. Do not imagine any contents that are not in the image.
}

Using the three aforementioned text generation strategies, we generate two textual descriptions per image. For images with attribute annotations (\emph{e.g.,} AG-ReID.v2~\cite{nguyen2024ag}, UAVHuman~\cite{li2021uav}), we combine attribute-based generation with either prompt-based or template-based generation, selected randomly. For images without attribute labels, we use a mix of prompt- and template-based generation strategies. This design ensures both the quality and diversity of textual descriptions.

\subsubsection{Test Set Correction.}
Given the inevitable noise in generated texts, we ensure the reliability of evaluation by employing six qualified annotators to manually inspect and revise the generated descriptions in the test set. This process enhances the accuracy and quality of the test data.

\subsection{Dataset Analysis}

\begin{figure}[t]
  \centering
  \includegraphics[width=0.9\linewidth, height=0.42\linewidth]{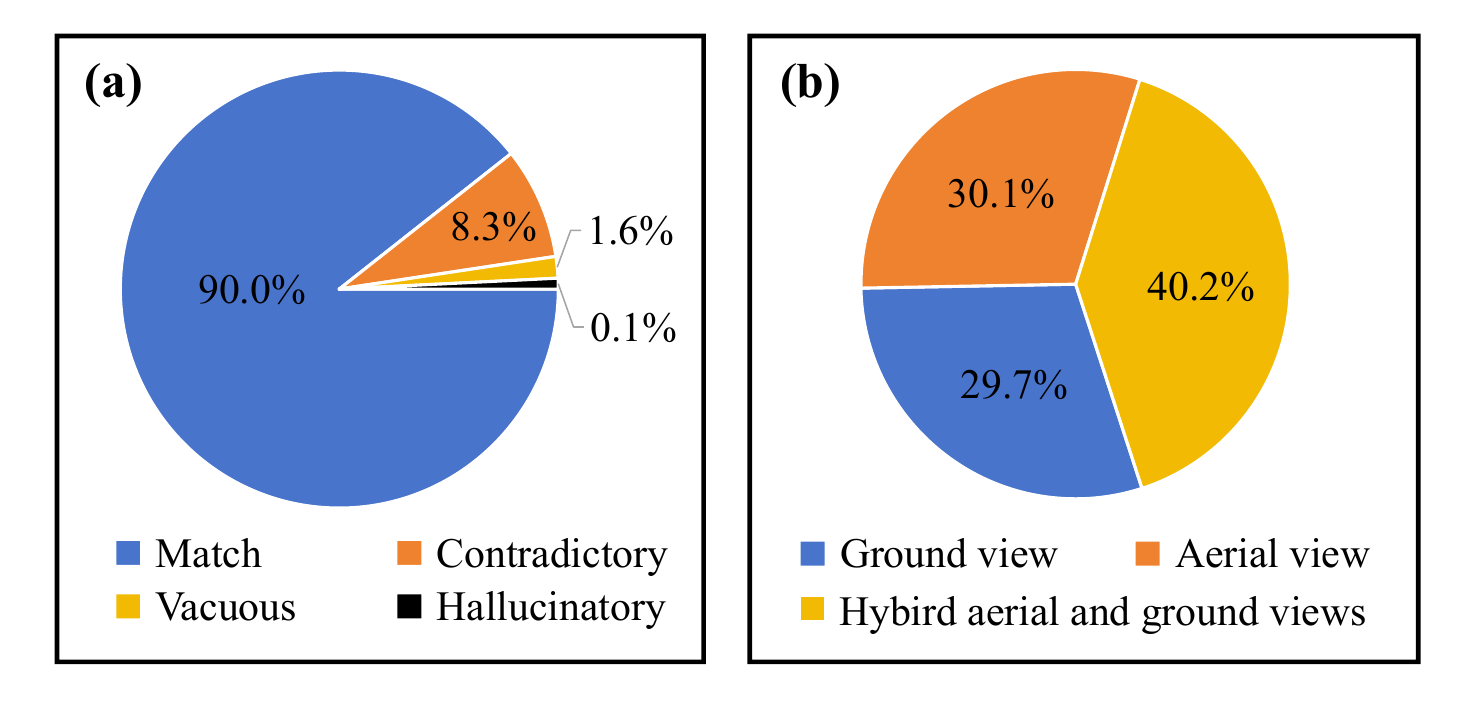}
  \caption{(a) Results of quality assessment on TAG-PEDES. (b) The proportion of identities from different viewpoints.}
  \label{fig: pie}
\end{figure}

We first conduct a quality assessment on TAG-PEDES to evaluate its reliability. To measure the alignment between generated descriptions and paired images, we define four evaluation categories: \textit{Match}, \textit{Contradictory}, \textit{Hallucinatory}, and \textit{Vacuous}. We use a crafted instruction to guide an MLLM~\cite{wang2024enhancing} for classification. As shown in Figure~\ref{fig: pie} (a), approximately $90\%$ of the descriptions are classified as \textit{Match}, indicating strong image-text alignment. Further details are provided in the Appendix.

We next highlight several distinctive characteristics of the proposed TAG-PEDES dataset. 
1) \textbf{Viewpoint Diversity.} Unlike traditional T-PR datasets that primarily contain  ground-view images, TAG-PEDES includes a balanced distribution of identities across diverse viewpoints, as shown in Figure~\ref{fig: pie} (b). 2) \textbf{High-Quality Synthetic Descriptions.} TAG-PEDES provides fine-grained textual descriptions averaging 39 words in length, generated by the proposed DTG and manually refined for test set. Compared to manual-only annotations in prior datasets, this ensures scalability. 3) \textbf{Increased Challenge.} The dataset poses difficulty by requiring to address both cross-modal alignment and multi-view heterogeneity, making it more challenging than existing Re-ID and T-PR benchmarks.
A comprehensive comparison with existing Re-ID and T-PR datasets is shown in the Appendix.

\section{Method}
\begin{figure*}[t]
  \centering
  \includegraphics[width=0.94\linewidth]{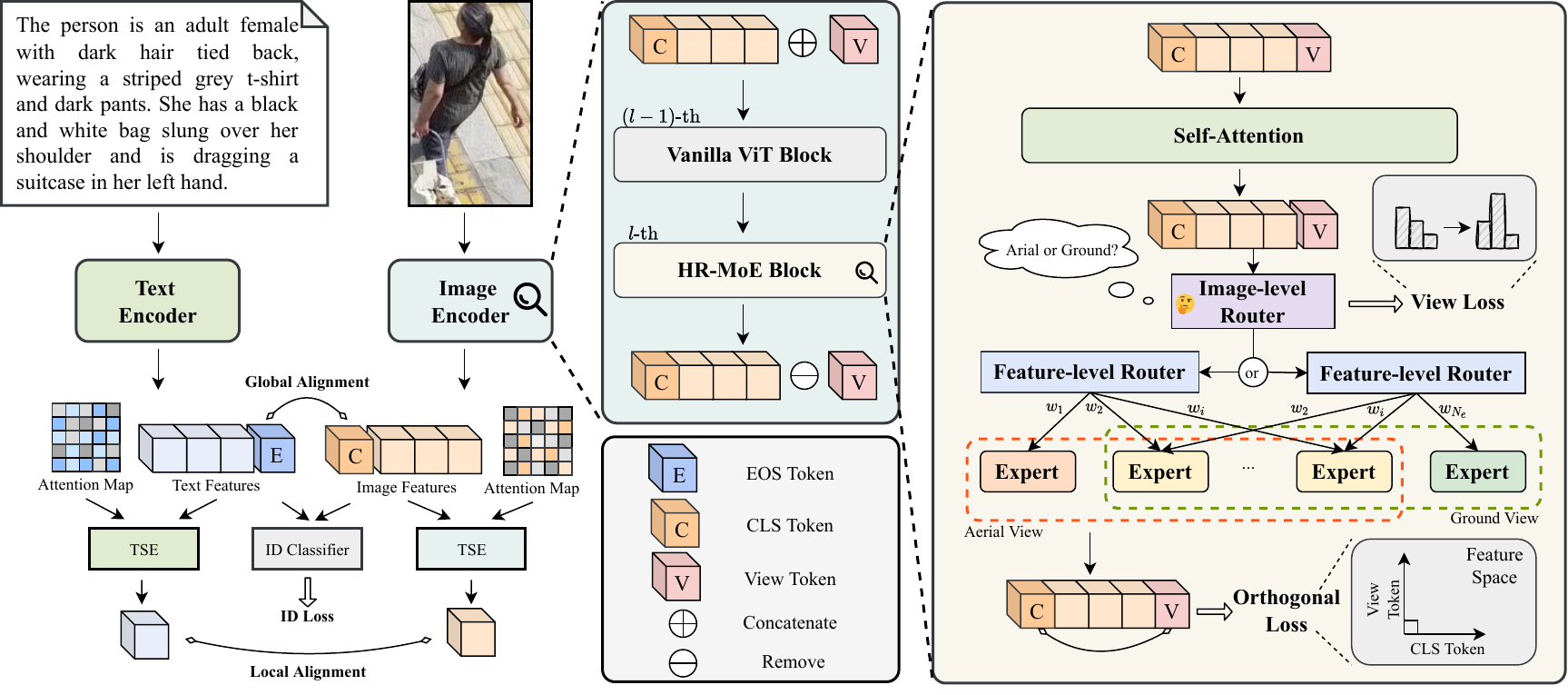}
  \caption{Overview of TAG-CLIP. It comprises an image encoder and a text encoder. Several ViT blocks in the image encoder are augmented with the HR-MoE module, which employs hierarchical routers and expert groups for robust visual feature extraction. A viewpoint decoupling strategy, consisting of two loss functions, is used to decouple viewpoint information from global visual features, thereby improving cross-modal alignment.}
  \label{fig: Method}
\end{figure*}
In this section, we elaborate on the proposed TAG-CLIP, as shown in Figure~\ref{fig: Method}. We begin with a brief review of the baseline model, followed by a detailed introduction of the proposed HR-MoE module and viewpoint decoupling strategy.

\subsection{Baseline}
We adopt TBPS-CLIP~\cite{cao2024empirical} as our baseline, a lightweight yet effective method for T-PR.
It consists of an image encoder and a text encoder to extract visual and textual features, respectively. The image encoder, consisting of $12$ Vision Transformer (ViT) blocks, takes an image $ I $ and divides it into $ S_I $ patches, which are linearly projected and combined with positional embeddings. 
A [CLS] token and a view token are prepended to the patch sequence before being fed into the encoder. The view token supports the HR-MoE module and viewpoint decoupling strategy detailed in the next section. This yields image feature \( F_v = \{v_{cls}, v_1, \dots, v_{S_I}, v_{view}\} \), where \( v_{cls} \) serves as the global visual feature, \(v_i\ (i=1,\cdots,S_I)\) are local features, and \(v_{view}\) is the view feature. Similarly, the input text is tokenized and augmented with [SOS] and [EOS] tokens to mark the start and end of text, and then passed through a Transformer-based text encoder to obtain textual features \( F_t = \{t_{sos}, t_1, \dots, t_{S_T}, t_{eos}\} \). \( t_{eos} \) serves as the global text feature and \(t_i\ (i=1,\cdots,S_T)\) are the local textual features.

The global alignment is then applied on global features:
\begin{equation}
    L_{GA} = L_{N-ITC}(v_{cls}, t_{eos}) + L_{R-ITC}(v_{cls}, t_{eos}).
\end{equation}
The loss \( L_{N-ITC}(\cdot) \) represents the alignment of image-text pairs by encouraging high similarity for positive pairs, while \( L_{R-ITC}(\cdot) \) focuses on pushing apart negative pairs by aligning the similarity with the label distribution in reverse. 
Further details are provided in ~\cite{cao2024empirical}.
In addition, we incorporate a Token Selection Embedding (TSE) module~\cite{qin2024noisy}, which selects and aggregates key local features, producing $ t_{tse}$ for text and $ v_{tse} $ for vision.
Then, local alignment which is performed as follows:
\begin{equation}
    L_{LA} = L_{N-ITC}(v_{tse}, t_{tse}) + L_{R-ITC}(v_{tse}, t_{tse}).
\end{equation}

Furthermore, we introduce an ID loss~\cite{zheng2020dual} \(L_{id}\), which enforces samples sharing the same identity to be clustered into a single class, promoting intra-class compactness and inter-class separability in the learned feature space.

\subsection{Hierarchically-Routed Mixture of Experts} 

To learn robust visual features from view-heterogeneous images, we introduce a Hierarchically-Routed Mixture of Experts (HR-MoE) module into a subset of ViT blocks in the image encoder, termed HR-MoE Blocks (see Figure~\ref{fig: Method}).

The HR-MoE Block processes image features \(F_v\) with a self-attention layer followed by the HR-MoE module.
The module consists of two hierarchical routers and a set of experts \(\{\mathcal{E}_i \mid i \in [1, N_e]\}\), which are divided into two groups specialized for aerial and ground views. The hierarchical routing mechanism allows each image feature to be dynamically routed to the most appropriate experts, enabling progressive learning of both view-specific and view-agnostic features from images captured under diverse viewpoints.

Specifically, the first router, referred to as the image level router \( \mathcal{R}_{Img} \), predicts the image's view category,
\begin{equation}\label{eq-g}
  g_{img} = \mathcal{R}_{Img}(v_{view}),  
\end{equation}
\begin{equation}
\label{eq: 4}
  z = \arg\max \left( Softmax(g_{img}) \right),
\end{equation}
where \( \mathcal{R}_{Img} \) is implemented as a linear classifier, \(g_{img}\) is the output one-hot logits, and \(z\) is the predicted binary view label (\(0\) for aerial, \(1\) for ground) and will be used to route the image features to the corresponding expert group in Eq.~\ref{eq-M} for subsequent specialized feature processing.

The second routing layer, termed the feature-level router \( \mathcal{R}_{Feat} \), performs a finer-grained assignment for each visual feature by selecting the most appropriate experts.
Formally, for the \( i \)-th visual feature \( F_v^i \) in the image features \( F_v \),
the routing process computes a probability distribution $P \in \mathbb{R}^{N_e}$ over all experts:
\begin{equation}
  g_{feat} = \mathcal{R}_{Feat}(F_v^i),
\end{equation}
\begin{equation}\label{eq-P}
  P = Softmax(g_{feat} + M).
\end{equation}
Here, $\mathcal{R}_{Feat}$ is implemented as a linear classification layer that outputs expert assignment logits \( g_{feat} \in \mathbb{R}^{N_e} \), where each element indicates the compatibility between feature $F_v^i$ and each expert.
\( M \in \mathbb{R}^{N_e} \), as a binary mask, is used to restrict expert selection based on the predicted view label $z$, obtained from the image-level router $\mathcal{R}_{Img}$.

Prior to defining $M$, we detail the expert group partitioning. The \( N_e \) experts is partitioned into two groups \( E_{aerial} = \{\mathcal{E}_n \mid n \in [1, e_0]\} \) for aerial-view images and \( E_{ground} = \{\mathcal{E}_n \mid n \in [e_1, N_e]\} \) for ground-view images.
Notably, \( e_1 < e_0 \), which implies a partial overlap between the two expert groups. 
The shared experts \( \{\mathcal{E}_n \mid n \in [e_1, e_0]\} \) are designed to capture view-agnostic features, while the non-overlapping ones in each group specialize in view-specific features.
Hereby, we define \(j\)-th element of the mask $M$,
\begin{equation}\label{eq-M}
    M_j =
\begin{cases}
0, & \text{if } z = 0 \text{ and } j \in [1, e_0], \\
0, & \text{if } z = 1 \text{ and } j \in [e_1, N_e], \\
-\infty, & \text{otherwise}.
\end{cases}
\end{equation}

This masking strategy enforces view-specific expert selection by setting certain entries in $g_{feat}$ to $-\infty$, forcing their softmax outputs to zero and disabling the corresponding experts. In contrast, entries set to $0$ remain unchanged, allowing the router to choose experts based on learned preferences within the permitted group.

Next, we perform a Top-\(K\) selection over the expert set for the $i$-th feature to identify the \( K \) most relevant experts according to the computed probability distribution $P$. Let \( \{w_{1}, \dots, w_{K}\} \) denote their corresponding normalized routing weights. 
The final updated features \(F^{i}_v \) of the $i$-th feature is obtained via a weighted summation of the expert outputs:
\begin{equation}
    F^{i}_v = \sum_{k=1}^{K} w_{k} \cdot \mathcal{E}_{k}(F^{i}_v),
\end{equation}
where $\mathcal{E}_{k}(\cdot)$ denotes the $k$-th expert. The resulting vector $F^{i}_v$ serves as the refined feature and is forwarded to the next layer for further processing.

\subsection{Viewpoint Decoupling Strategy}
To perform effective cross-modal alignment, it is critical to address the inherent asymmetry between visual and textual features, that is, image features contain explicit viewpoint information, whereas textual descriptions typically lack such cues. In response, we propose a viewpoint decoupling strategy, comprising a view loss and an orthogonal loss.

The view loss trains the view feature $v_{view}$ to explicitly capture viewpoint information from images. Formally, we minimize the cross-entropy between the logits $g_{img} $ in Eq.~\ref{eq-g} and the one-hot ground-truth view label vector $z_{gt}$:
\begin{equation}
L_{view} = \mathcal{H}(g_{img}, z_{gt}),
\end{equation}
where \(\mathcal{H}\) represents the cross-entropy computation.

The orthogonal loss is used to decouple viewpoint-specific information from the global feature $v_{cls}$ by enforcing orthogonality between the view feature $v_{view}$ and the global feature $v_{cls}$ in feature space.
The loss is defined as:
\begin{equation}
\label{eq: 10}
L_{ortho} = \min\left(|\cos(v_{cls}, v_{view})|, \alpha\right),
\end{equation}
where $|\cdot|$ denotes the absolute value, $\cos$ represents cosine similarity, and $\alpha$ is a predefined threshold.
This geometric constraint encourages orthogonality between the feature vectors to effectively decouple most viewpoint-specific information from $v_{cls}$. The threshold $\alpha$, in turn, prevents the exclusion of viewpoint-specific yet highly discriminative pedestrian features from $v_{cls}$. As a result, the global visual feature $v_{cls}$ retains both viewpoint agnosticism and discriminability, making it well-suited for cross-modal alignment. 

\subsection{Training and Inference}
In the training phase, we jointly optimize both alignment-oriented and viewpoint-decoupling loss functions:
\begin{equation}
    L = L_1 + L_2,
\end{equation}
\begin{equation} \label{eq: 15}
    L_1 = L_{GA} + L_{LA} + \lambda_{id} L_{id},
\end{equation}
\begin{equation} \label{eq: 16}
    L_2 = L_{view} + \lambda_{ortho} L_{ortho}.
\end{equation}
Here, \( \lambda_{id} \) and \( \lambda_{ortho} \) serve as hyperparameters.

In inference, the cross-modal similarity score \( sim \) is computed as the average of global and local alignment scores:  
\begin{equation}
    sim = \frac{1}{2} [\cos(v_{cls}, t_{eos}) + \cos(v_{tse}, t_{tse})].
\end{equation}

\FloatBarrier
\section{Experiments}
We evaluate our method on four datasets: our newly proposed TAG-PEDES, along with three widely used T-PR datasets CUHK-PEDES~\cite{li2017identity}, ICFG-PEDES~\cite{ding2021semantically} and RSTPReID~\cite{zhu2021dssl}.

\textbf{TAG-PEDES} has person images captured from diverse viewpoints, comprising $28,206$ images of $6,840$ identities, each paired with $2$ textual descriptions. The dataset is divided into a training set and a test set. The training set consists of $19,954$ images from $4,840$ identities and $39,908$ text descriptions. The test set includes $8,252$ images, $16,50
4$ text descriptions, and $2,000$ identities.

Details on the T-PR datasets, evaluation metrics, and implementation are provided in the Appendix.

\begin{table}[t]
    \centering
    \fontsize{9}{11}\selectfont
    \setlength{\tabcolsep}{1mm}
    \begin{tabular}{l|cccc}
    \hline

    Methods                           & R@1   & R@5   & R@10  & mAP    \\ 
    \hline\hline
    \multicolumn{5}{l}{\textit{w/o CLIP:}} \\
    \hline
    RaSa~\cite{bai2023rasa}           & 61.32 & 78.31 & 84.03 & 48.77 \\
    APTM~\cite{yang2023towards}       & 61.52 & 78.25 & 83.94 & 47.77 \\
    AUL~\cite{li2024adaptive}         & 59.26 & 76.64 & 82.89 & 45.79  \\
    \hline
    \multicolumn{5}{l}{\textit{w/ CLIP:}} \\
    \hline
    CFine~\cite{yan2023clip}         & 56.68 & 74.78 & 81.34 & - \\
    IRRA~\cite{jiang2023cross}       & 59.61 & 77.52 & 83.56 & 46.53 \\
    TBPS-CLIP~\cite{cao2024empirical}  & 60.28 & 77.71 & 83.70 & 46.45 \\
    RDE~\cite{qin2024noisy}          & 61.58 & 78.54 & 84.08 & 48.03 \\
    AEA-FIRM~\cite{wang2025aea}     & 51.83 & 71.78 & 79.01 & 38.60 \\
    TAG-CLIP (Ours)                        & \textbf{62.64} & \textbf{79.10} & \textbf{84.81} & \textbf{49.27} \\ 
    \hline
    \end{tabular}
    \caption{Comparison with state-of-the-art methods on TAG-PEDES. We reproduce these compared methods with their official released code.}
    \label{tab: t2ag}
\end{table}

\begin{table}[t]
    \fontsize{9}{11}\selectfont
    \setlength{\tabcolsep}{1mm}
    \centering

    \begin{tabular}{l|cccc}
    \hline
 
    Methods       & R@1   & R@5   & R@10   & mAP  \\ 
    \hline\hline
    \multicolumn{5}{l}{\textit{w/o CLIP:}} \\
    \hline
    LGUR~\cite{shao2022learning}                     & 65.25 & 83.12 & 89.00 &   -   \\ 
    RaSa~\cite{bai2023rasa}                 & 76.51 & 90.29 & 94.25 & \textbf{69.38} \\ 
    APTM~\cite{yang2023towards}             & 76.53 & 90.04 & 94.15 & 66.91 \\ 
    AUL~\cite{li2024adaptive}               & \textbf{77.23} & \textbf{90.43} & \textbf{94.41} &   -  \\ 
    \hline
    \multicolumn{5}{l}{\textit{w/ CLIP:}} \\
    \hline
    CFine~\cite{yan2023clip}                & 69.57 & 85.93 & 91.15 &   - \\
    IRRA~\cite{jiang2023cross}              & 73.38 & 89.93 & 93.71 & 66.13 \\
    Propot~\cite{yan2024prototypical}       & 74.89 & 89.90 & 94.17 & 67.12 \\ 
    TBPS-CLIP~\cite{cao2024empirical}       & 72.66 & 88.14 & 92.72 & 64.97 \\ 
    RDE~\cite{qin2024noisy}                 & 75.94 & 90.14 & 94.12 & 67.56  \\ 
    TAG-CLIP (Ours)                              & 74.38 & 88.30 & 92.59 & 67.18 \\
    \hline
    \end{tabular}
    \caption{Comparison with state-of-the-art methods on CUHK-PEDES.}
    \label{tab: t2g}
\end{table}

\subsection{Comparison with State-of-the-Art Methods}
We evaluate the proposed method on the newly constructed TAG-PEDES dataset, aiming to evaluate its effectiveness.
As shown in Table~\ref{tab: t2ag}, we compare TAG-CLIP with several representative T-PR methods, as well as the concurrent TAG-PR method AEA-FIRM. All competing methods are reproduced on TAG-PEDES dataset using their publicly available code.
Firstly, our method outperforms all T-PR methods, regardless of whether they are based on CLIP, achieving the best performance on both R@1 and mAP metrics. Specifically, TAG-CLIP surpasses the state-of-the-art method RDE~\cite{qin2024noisy} by $1.06\%$ and $1.24\%$ in R@1 and mAP, respectively.
Secondly, compared to AEA-FIRM, which is specialized for TAG-PR, our method demonstrates superior performance. AEA-FIRM employs an AEA loss with a strict triplet input structure, which limits its adaptability to view variations, whereas our TAG-CLIP explicitly models view heterogeneity during both image encoding and cross-modal alignment, resulting in better performance.

In addition, we evaluate TAG-CLIP on the conventional ground-only dataset CUHK-PEDES and compare it with state-of-the-art methods, as shown in Table~\ref{tab: t2g}. \textbf{It is reasonable that TAG-CLIP exhibits slightly inferior performance in this setting, due to the following factors.} First, under the single-view setting where all images are captured from the ground viewpoint, both the HR-MoE module and the viewpoint decoupling strategy in TAG-CLIP become inapplicable. Consequently, TAG-CLIP essentially degenerates into a baseline model, resulting in suboptimal performance compared to methods \textbf{\emph{specialized for single-view scenarios}}. Second, our dual-stream CLIP-based architecture lacks the advantage of cross-modal fusion modules leveraged by one-stream models such as AUL~\cite{li2024adaptive}, which further contributes to the performance gap.

In the Appendix, we present evaluations of TAG-CLIP on ground-only datasets (ICFG-PEDES and RSTPReid) and aerial-only setting to further demonstrate its effectiveness.

\subsection{Ablation Study}
In this section, we perform ablation studies to evaluate the effectiveness of the proposed HR-MoE module and viewpoint decoupling strategy in TAG-CLIP.

\begin{figure}[tbh]
  \centering
  \includegraphics[width=1\linewidth]{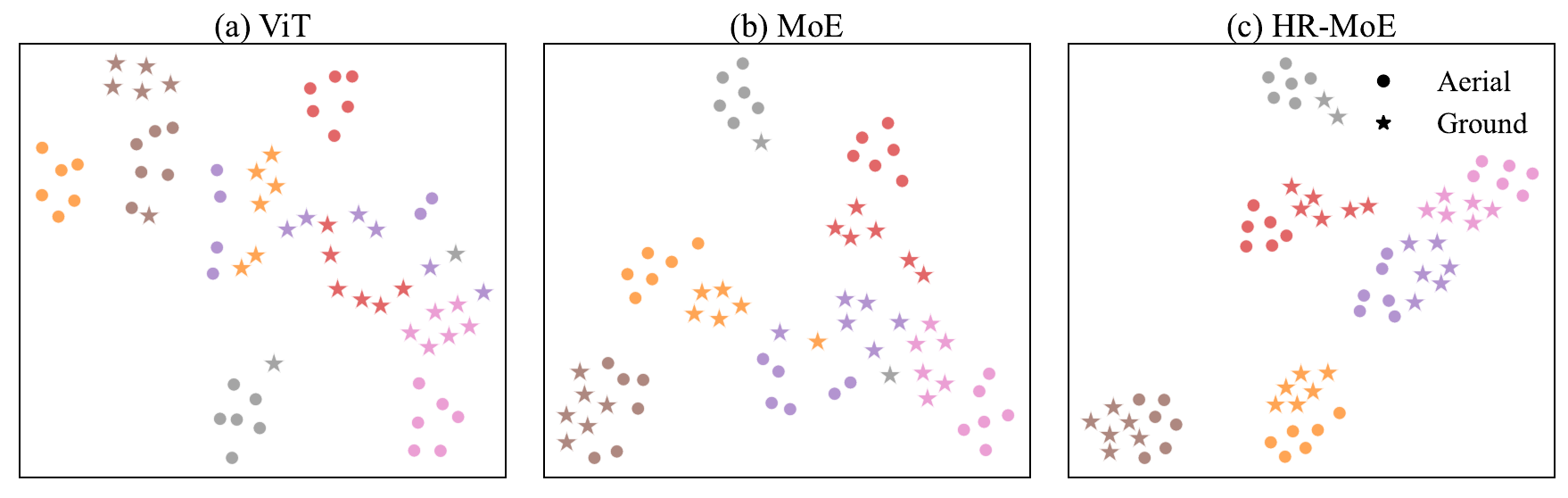}
  \caption{t-SNE visualization of extracted visual features. Each color represents a unique identity.}
  \label{fig: tsne}
\end{figure}

\textbf{Ablation on HR-MoE.}
We propose HR-MoE, integrated into the image encoder to extract robust visual features from heterogeneous image views. To verify its effectiveness, we compare the HR-MoE block with two alternative configurations: the standard ViT block and the vanilla MoE structure. The former refers to the baseline where HR-MoE is not applied to the image encoder, while the latter incorporates a set of experts for feature learning without employing the hierarchical routing mechanism.
The results are summarized in Table~\ref{tab: ablation_study}. 
First, when comparing the ViT block with HR-MoE (No.1 \emph{vs.} No.3), we observe performance gains, demonstrating the effectiveness of the HR-MoE design. 
Second, in contrast to the vanilla MoE (No.2 \emph{vs.} No.3), HR-MoE achieves a more substantial improvement. This result suggests that our hierarchical routing strategy and expert grouping mechanism are critical to achieving superior performance.
We further visualize features from ViT, MoE, and HR-MoE with t-SNE (Fig.~\ref{fig: tsne}), showing that HR-MoE yields more discriminative features with reduced viewpoint sensitivity.

\begin{table}[tb]
    \fontsize{9}{11}\selectfont
    \setlength{\tabcolsep}{0.5mm}
    \centering


    \begin{tabular}{c|ccc|cc|cccc}
    \hline

      &\multicolumn{3}{c|}{Block} & \multicolumn{2}{c|}{Decoupling}  &     &    &     &      \\ 

    \multirow{-2}{*}{No.}  & ViT & MoE & HR-MoE & \(L_{view}\) & \(L_{ortho}\)  & \multirow{-2}{*}{R@1}   & \multirow{-2}{*}{R@5}   & \multirow{-2}{*}{R@10}   & \multirow{-2}{*}{mAP}    \\ 
    \hline\hline
    1            & \ding{51} &            &            &            &           & 59.09 & 76.58 & 82.55 & 48.05   \\
    2            &            & \ding{51} &            &            &           & 60.26 & 78.01 & 84.06 & 48.95   \\
    3            &            &            & \ding{51} &            &           & 61.60 & 78.45 & 84.39 & 48.94   \\
    \hline
    4            &            &            & \ding{51} & \ding{51} &            & 61.55 & 78.41 & 84.08 & 48.73   \\ 
    5            &            &            & \ding{51} &            & \ding{51} & 62.32 & \textbf{79.37} & 84.51 & 49.22   \\ 
    \hline
    6            &            &            & \ding{51} & \ding{51} & \ding{51} & \textbf{62.64} & 79.10 & \textbf{84.81} & \textbf{49.27}  \\ 
    
    \hline
    \end{tabular}
    \caption{Ablation results of TAG-CLIP's components.}
    \label{tab: ablation_study}
\end{table}

\begin{table}[tb]
    \centering
    \fontsize{9}{11}\selectfont
    \setlength{\tabcolsep}{0.6mm}
    
    \begin{tabular}{c|cc|cccc}
    \hline

    No. & Train Set & Test Set & R@1    & R@5    & R@10   & mAP    \\ 
    \hline\hline
    1 & Aerial          & Aerial         & 54.04 & 73.74 & 80.61 & 51.45 \\
    2 & Aerial\&Ground  & Aerial         & 54.64 & 73.01 & 80.40 & 51.77 \\
    \hline
    3 & Ground          & Ground         & 63.57 & 80.93 & 86.29 & 60.31 \\
    4 & Aerial\&Ground  & Ground         & 64.49 & 80.57 & 86.55 & 60.76 \\
    \hline
    5 & Aerial          & Aerial\&Ground & 47.96 & 69.48 & 76.76 & 36.02 \\
    6 & Ground          & Aerial\&Ground & 53.48 & 73.03 & 79.44 & 41.19 \\
    7 & Aerial\&Ground  & Aerial\&Ground & 57.80 & 76.90 & 82.58 & 44.40 \\

    \hline
    \end{tabular}
    \caption{Ablation results of diverse viewpoints.}
    \label{tab: viewpoint}
\end{table}

\begin{figure}[tbh]
  \centering
  \includegraphics[width=0.92\linewidth, height=0.38\linewidth]{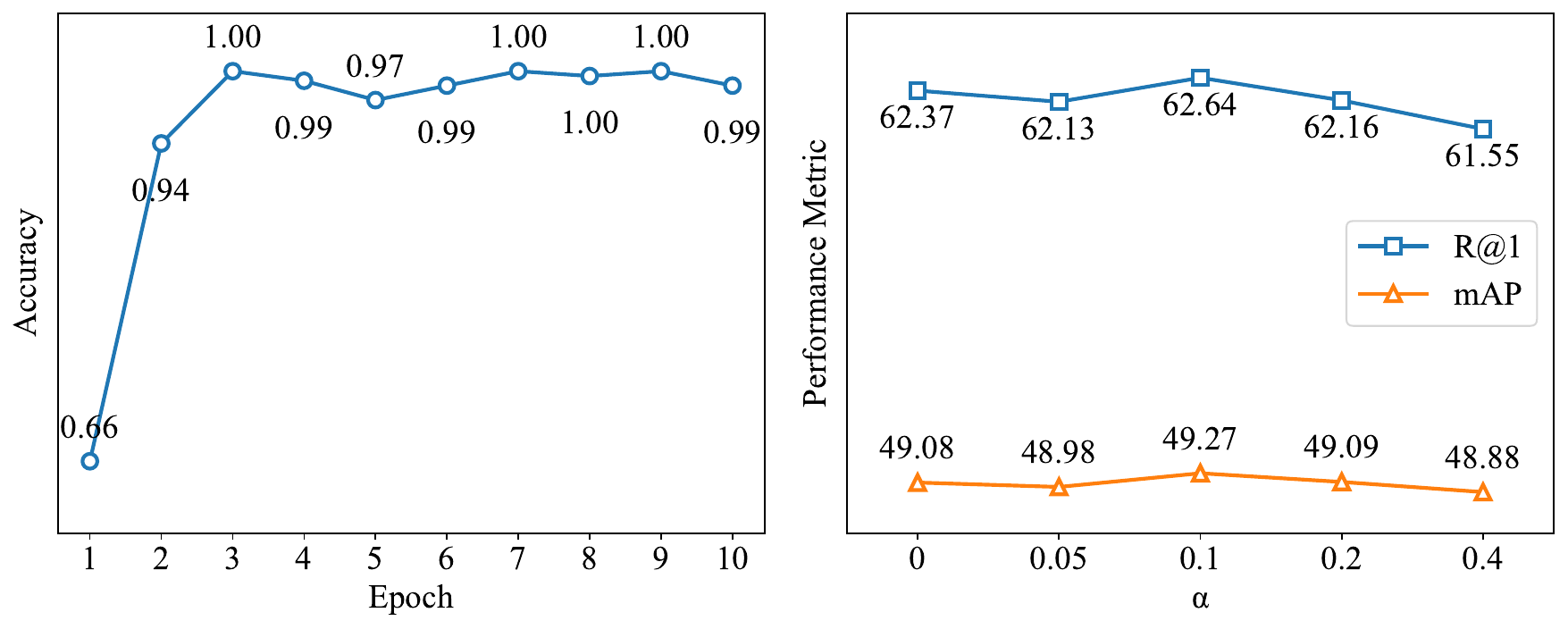}
  \caption{Left: Viewpoint classification accuracy of Image-level Router (Eq.~\ref{eq: 4}). Right: Hyperparameter analysis of $\alpha$.}
  \label{fig: hyperpara}
\end{figure}

\textbf{Effectiveness of view decoupling strategy.}
In the proposed view decoupling strategy, the view loss and orthogonal loss jointly facilitate viewpoint decoupling: the view loss explicitly guides the view token to capture viewpoint-specific features, while the orthogonal loss minimizes viewpoint information in the global feature.
To assess their effectiveness, we conduct ablation studies by individually removing each loss, as shown in Table~\ref{tab: ablation_study}. 
It can be observed that (1) removing the orthogonal loss results in performance drop (No.4 \emph{vs.} No.6), highlighting the negative effect of retaining viewpoint-specific information during image feature extraction and subsequent cross-modal alignment; (2) removing the view loss leads to a slight performance drop (No.5 \emph{vs.} No.6), indicating that the model can still capture certain viewpoint-specific features even without explicit supervision. Additionally, Fig.~\ref{fig: hyperpara} (left) shows that, the image-level router achieves high viewpoint classification accuracy, under the supervision of the view loss.

\textbf{Ablation on viewpoints.}
To investigate the impact of diverse viewpoints, we evaluate TAG-CLIP on different viewpoint settings. Specifically, we select all $2,747$ identities with multi-view images from TAG-PEDES and randomly split them into a two groups (\emph{i.e.}, $2,000$ IDs for training and $747$ IDs for testing). Based on this, we construct three sub-datasets: one with only aerial-view images, one with only ground-view images, and one with a mix of aerial and ground views. All three datasets contain the same set of identities, differing only in the viewpoint of images.
We then report the retrieval performance in Table~\ref{tab: viewpoint}. TAG-CLIP trained on multi-view data consistently outperforms its single-view counterparts across all evaluation settings (No.1 \emph{vs.} No.2, No.3 \emph{vs.} No.4, and No.5/No.6 \emph{vs.} No.7), which stems from two factors: (1) the model’s design effectively leverages multi-view information, and (2) multi-view training enhances feature robustness and generalization.

\subsection{Hyperparameter Analysis}
The hyperparameter $\alpha$ in Eq.~\ref{eq: 10} controls the amount of viewpoint-specific information retained in the global visual feature. As shown in Fig.~\ref{fig: hyperpara} (right), performance drops when $\alpha$ is too large (over-preserving view-specific cues) or too small (losing discriminative signals). At $\alpha=0$, $L_{ortho}$ becomes a standard orthogonal loss, which overly suppresses useful features. We set $\alpha=0.1$ for trade-off.
The analysis of $\lambda_{id}$ and $\lambda_{ortho}$ is provided in the Appendix.

\section{Conclusion}
In this paper, we introduce TAG-PR, which tackles person retrieval across a hybrid of aerial and ground image views, better reflecting real-world scenarios. To facilitate research in this task, we construct TAG-PEDES, which features diverse-view person images and high-quality synthetic textual descriptions.
To address the challenges of heterogeneous image views, we propose TAG-CLIP, a method that incorporates the HR-MoE module and viewpoint decoupling strategy to enhance both image feature learning and cross-modal alignment.
Extensive experiments demonstrate the effectiveness of TAG-CLIP.
It is expected that our work can advance the field of person retrieval and drive its practical application in real-world scenarios.

\bibliography{aaai2026}

\section{More Dataset Details}
\subsection{More Examples from TAG-PEDES}
\begin{figure*}[tb]
  \centering
  \includegraphics[width=\linewidth]{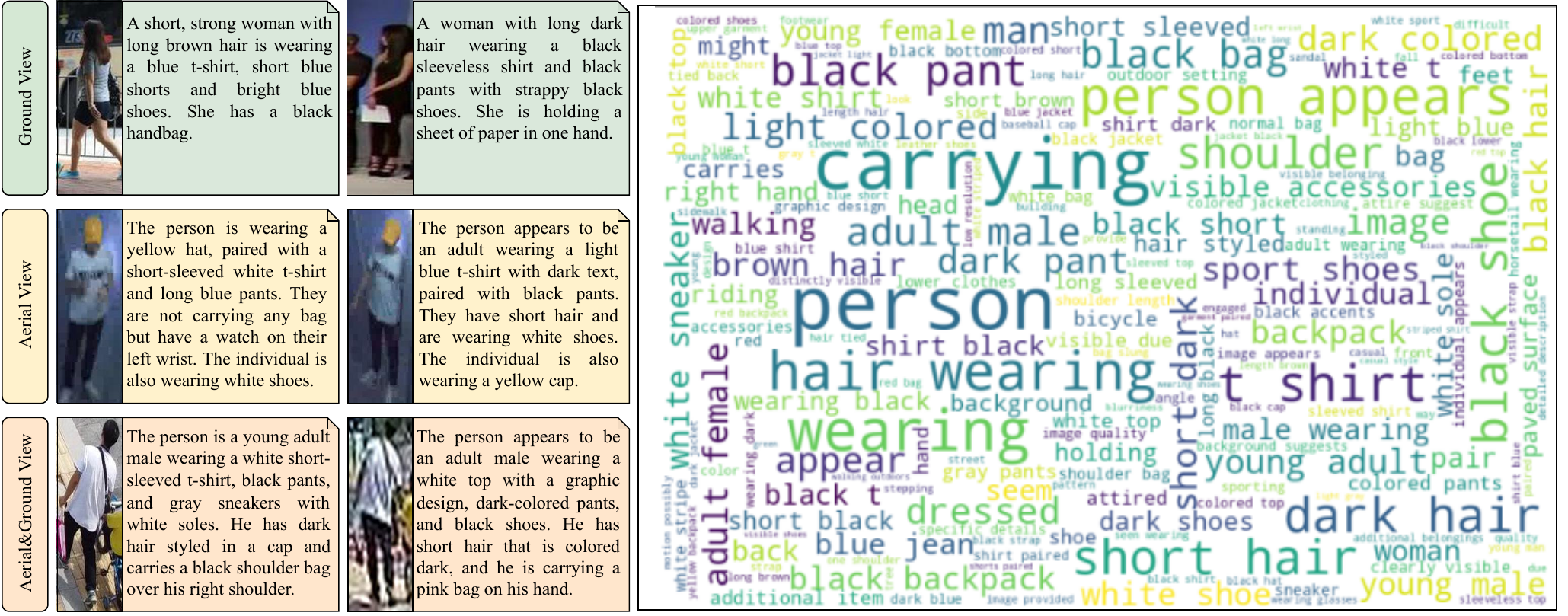}
  \caption{Some visualization examples from our proposed TAG-PEDES and a word cloud of this dataset.}
  \label{fig: imgexample}
\end{figure*}

Figure~\ref{fig: imgexample} showcases more examples from  TAG-PEDES dataset along with a word cloud visualization. The dataset contains pedestrian images captured from diverse viewpoints. Ground-view images are typically taken at eye level, clearly presenting full-body attributes such as hair, upper clothing, pants, and shoes. In contrast, aerial-view images are captured from significantly higher angles, often leading to occlusion of lower-body attributes—especially features like footwear. Even for the same identity, images from different viewpoints can vary drastically. Compared to traditional T-PR task, TAG-PEDES better reflects the complexity and diversity of real-world scenarios, making it more challenging and practical.

\begin{table*}[tb]
    \centering
    \fontsize{9}{11}\selectfont
    \setlength{\tabcolsep}{1mm}
    \begin{tabular}{ccc|cccc}
    \hline

    Task & Modality & Dataset    & \#IDs    &   \#Images       & \#Texts  & View  \\ 
    \hline\hline
    & & Market-1501~\cite{zheng2015person}  & $1,501$  & $32,668$ & -  & Ground   \\
    & & PRAI-1581~\cite{zhang2020person}    & $1,581$  & $39,461$ & -  & Aerial   \\
    \multirow{-3}{*}{Re-ID} & \multirow{-3}{*}{Image} & AG-ReID.v2~\cite{nguyen2024ag}        & $1,615$ & $100,502$ & - & Aerial\&Ground \\
    \hline 
    & & CUHK-PEDES~\cite{li2017person}        & $13,003$ & $40,206$ & $80,440$ & Ground  \\
    & & ICFG-PEDES~\cite{ding2021semantically} & $4,102$  & $54,522$ & $54,522$ & Ground   \\
    & & RSTPReid~\cite{zhu2021dssl}           & $4,101$  & $20,505$ & $41,010$ & Ground   \\
    \multirow{-4}{*}{T-PR} & \multirow{-4}{*}{Text-Image} & UFine6926~\cite{zuo2024ufinebench}     & $6,926$  & $26,206$ & $52,412$ & Ground   \\
    \hline
    TAG-PR &  Text-Image & TAG-PEDES                              & $6,840$  & $28,206$  & $56,412$ & Aerial\&Ground \\
    \hline
    \end{tabular}
    \caption{Comparison between our TAG-PEDES dataset and existing Re-ID and T-PR datasets. “\#” denotes the number of instances, and “Avg. Text” refers to the average text length.}
    \label{tab: Com2ExiDatasets}
\end{table*}

\subsection{Comparsion with Existing ReID and T-PR Datasets}
In Table~\ref{tab: Com2ExiDatasets}, we compare TAG-PEDES with existing person re-identification (ReID) and text-based person retrieval (T-PR) datasets.
ReID datasets typically contain only the visual modality. Although some incorporate aerial-ground viewpoint discrepancies, they are still confined to unimodal settings, relying solely on visual information for person retrieval. Existing T-PR datasets provide paired image-text data, introducing greater retrieval difficulty. However, these datasets are predominantly composed of ground-view images, lacking the viewpoint diversity and complexity in real-world scenarios, which involve heterogeneous viewpoints.

In contrast, our TAG-PEDES dataset is specifically designed to bridge this gap. It incorporates person images captured from both aerial and ground views and provides high-quality textual descriptions for each image, introducing a higher level of difficulty and better reflects real-world scenarios. Therefore, TAG-PEDES serves as a more comprehensive and challenging benchmark for advancing research in robust and practical person retrieval.

\subsection{Details of Templates}
To ensure the diversity of generated texts, we employed a large set of templates in both template-based and attribute-based generation strategies. Below, we detail the process of obtaining these templates. 

\begin{figure*}[tbh]
  \centering
  \includegraphics[width=\linewidth]{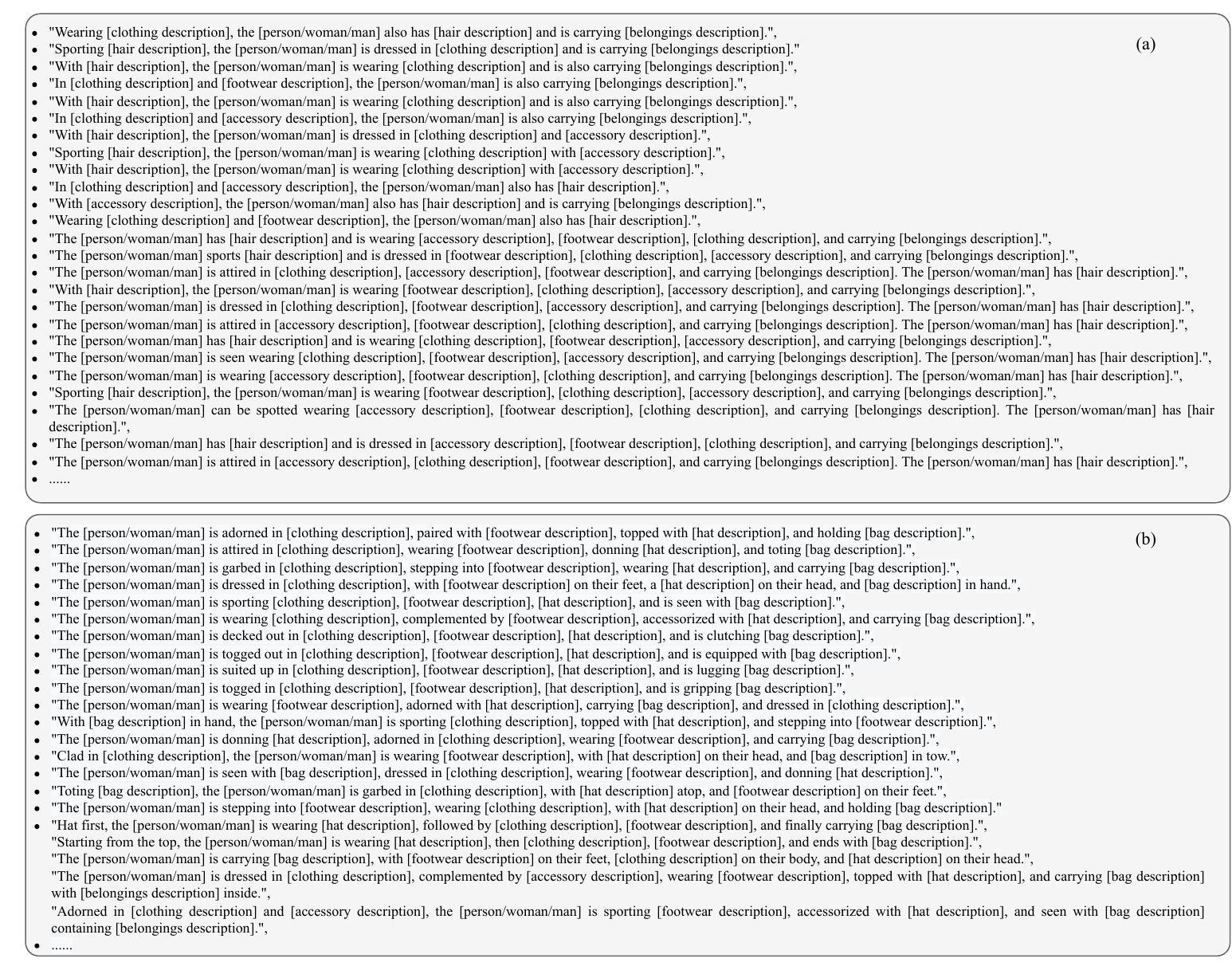}
  \caption{Some template examples from the template-based and attribute-based generation strategies.}
  \label{fig: temps}
\end{figure*}

Starting with \(46\) templates from \cite{tan2024harnessing}, we manually evaluate small batches of generated samples based on accuracy, and ultimately retain \(31\) reliable templates for the template-based generation strategy. Figure~\ref{fig: temps} (a) shows several representative templates.

For the attribute-based generation method, since it relies on explicit attribute labels, mismatches between the provided labels and the descriptions required by generic templates can exacerbate hallucination issues in MLLMs. Therefore, we customize the templates to align with the available attributes. For instance, in the UAVHuman~\cite{li2021uav} dataset, which provides seven attribute labels (\emph{e.g.} gender, hat, backpack, upper clothing color and style, and lower clothing color and style), we customize the template as follows:

\textit{The [person/woman/man] is adorned in [clothing description], paired with [footwear description], topped with [hat description], and holding [bag description]. 
}

This ensures consistency between the template and the attribute labels. Following this principle, we derive \(28\) additional templates tailored for attribute-based generation from the \(31\) original templates, shown in Figure~\ref{fig: temps} (b). 
These two template pools effectively support their respective methods in generating diverse and reliable textual descriptions.

\subsection{Details of Quality Assessment}
\begin{figure*}[tbh]
  \centering
  \includegraphics[width=\linewidth]{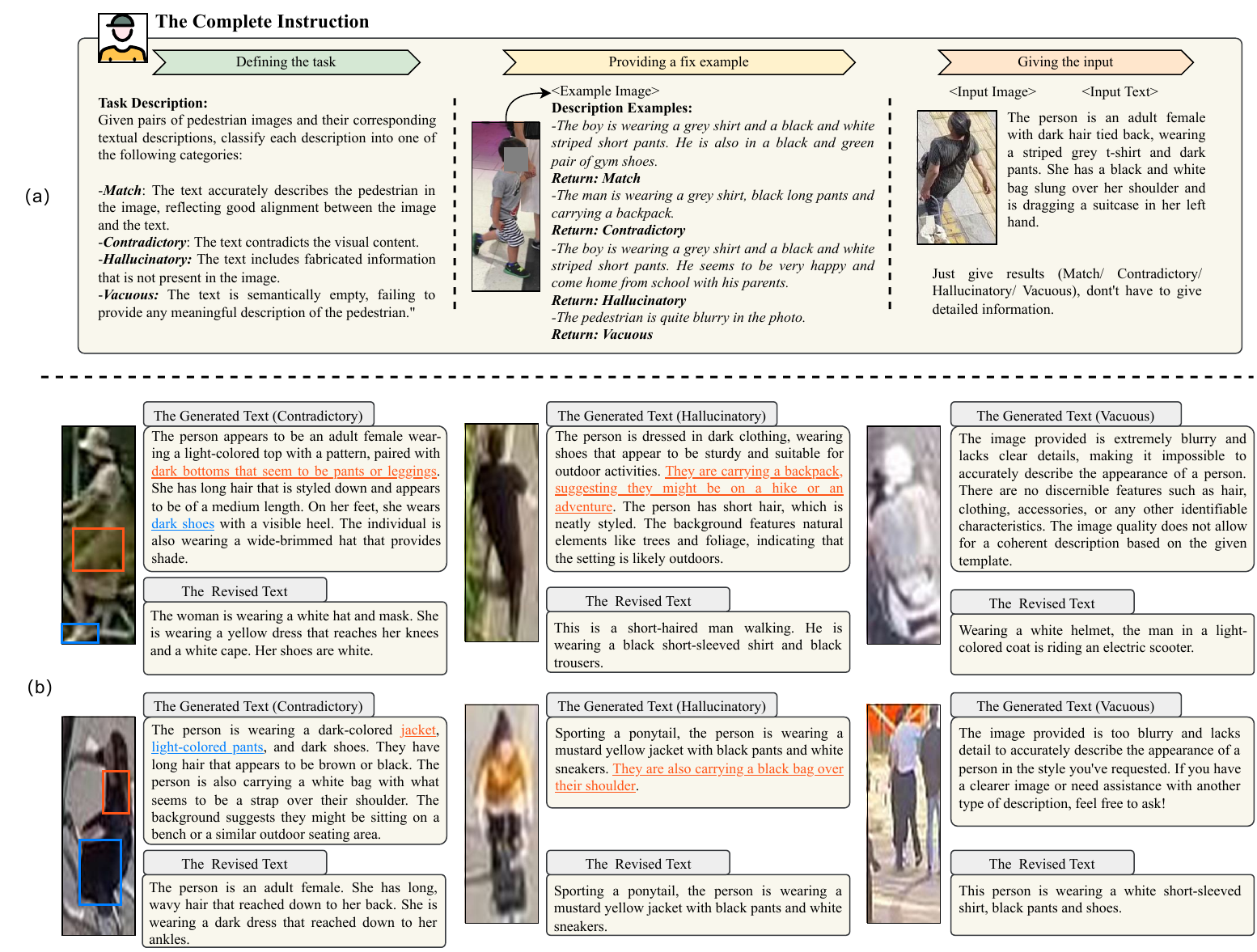}
  \caption{(a) Instruction used for MLLM evaluation. The instruction consists of three parts: defining the task, providing a fix example, and presenting the input image-text pair. This structure enables the MLLM to fully understand the evaluation task and accurately assess the quality of the generated texts. (b) Visualization of the evaluation results from the test set produced by the MLLM. Each column presents a specific example of one mismatch type, along with its manually revised text. Different colors are used to highlight errors in the text or inconsistencies between the text and the image.}
  \label{fig: instruction&examples}
\end{figure*}

We design a crafted instruction to prompt the MLLM~
\cite{wang2024enhancing} for accurate evaluation of the generated text. As illustrated in Figure~\ref{fig: instruction&examples} (a), the instruction consists of three parts. First, we provide a task definition, asking the MLLM to classify each input image-text pair into one of four predefined categories: Match, Contradictory, Hallucinatory, and Vacuous:
\begin{itemize}
    \item \textbf{Matched}: The text accurately describes the pedestrian in the image, reflecting strong cross-modal alignment.  
    \item \textbf{Contradictory}: The text presents information that directly conflicts with the visual content of the image. 
    \item \textbf{Hallucinatory}: The text includes fabricated information that is not present in the image.  
    \item \textbf{Vacuous}: The text is semantically empty, failing to provide any meaningful description of the pedestrian.
\end{itemize} 
To help the MLLM better understand the task, we then present a concrete example, including an image and four associated text descriptions, each annotated with its expected category. Finally, the actual image-text pairs are provided as input for classification.

Figure~\ref{fig: instruction&examples} (b) presents several evaluation results from the test set, along with the corresponding revised texts for comparison. As shown, when MLLMs are faced with blurred or low-resolution images, the generated texts often contain errors, which typically fall into three categories: (1) incorrect descriptions of pedestrian attributes, such as misidentifying a green dress as black pants; (2) hallucinated content, where nonexistent objects are mentioned; and (3) failure to provide any meaningful information. These errors can be corrected through subsequent human review and refinement, thereby ensuring the reliability of the data used for model performance evaluation.

\section{Further Experiments}
\subsection{Datasets and Evaluation Metrics}
\textbf{CUHK-PEDES}~\cite{li2017person} is the most widely used dataset for T-PR, containing \( 40,206 \) ground-veiw images and \( 80,440 \) textual descriptions from \( 13,003 \) unique identities. The dataset is officially split into \( 34,054 \) training images with \( 68,126 \) texts from \( 11,003 \) identities, \( 3,078 \) validation images with \( 6,158 \) texts from \( 1,000 \) identities, and \( 3,074 \) test images with \( 6,156 \) texts from \( 1,000 \) identities.

\textbf{ICFG-PEDES}~\cite{ding2021semantically} consists of \( 54,522 \) ground-view images from \( 4,102 \) identities, each paired with a textual description averaging \( 37 \) words. It is divided into \( 34,674 \) training images from \( 3,102 \) identities and \( 19,848 \) test images from \( 1,000 \) identities.

\textbf{RSTPReID}~\cite{zhu2021dssl} focuses on real-world surveillance scenarios, comprising \( 20,505 \) ground-view images from \( 4,101 \) identities with five images per identity. Each image is annotated with two detailed textual descriptions of at least \( 23 \) words. The dataset is split into \( 3,701 \) training identities, \( 200 \) validation identities, and \( 200 \) test identities.

\textbf{Evaluation Metrics}. We use the popular Rank@$k$ (R@$k$ for short, $k = 1,5,10$) to evaluate the performance of the methods. 
The mean Average Precision (mAP) is also adopted for a comprehensive evaluation. The higher R@$k$ and mAP indicate better performance.

\subsection{Implementation Details}
For data generation, we employ LLaVA-OneVision~\cite{li2024llava} as the generation model and InternVL2.5-MPO~\cite{wang2024enhancing} as the evaluation model.
The proposed HR-MoE module is integrated into the \(7\)th to \(9\)th ViT blocks within the image encoder of the baseline model. The total number of experts is set to \( N_e = 6 \), with \( e_0 = 5 \) and \( e_1 = 1 \), meaning each expert group includes \(1\) view-specific expert and \(4\) shared experts. For each visual token, the top-\(5\) experts are selected based on the routing probabilities \( P \).
The hyperparameters in Eq.~10, Eq.~15 and Eq.~16 are set as \( \alpha = 0.1 \), \( \lambda_{id} = 0.5 \) and \( \lambda_{ortho} = 100 \), respectively. During training, input images are resized to \(224 \times 224\), and the maximum text length is set to \(77\). We use the AdamW optimizer with linear warm-up and cosine decay scheduling, starting from an initial learning rate of \(1 \times 10^{-6}\), peaking at \(1 \times 10^{-4}\), and decaying to \(5 \times 10^{-6}\). The model is trained for \(10\) epochs with batchsize \(= 80\). All experiments are conducted on four NVIDIA 3090 GPUs.

\subsection{Comparisons with State-of-the-Art Methods}
We conduct additional experiments to evaluate the performance of the proposed TAG-CLIP on both text-based aerial person retrieval and traditional T-PR datasets.
\emph{In these tasks, all images are captured from a single viewpoint—either aerial-only (for text-based aerial person retrieval) or ground-only (for traditional T-PR). As a result, the proposed HR-MoE module and the viewpoint decoupling strategy in TAG-CLIP become inapplicable, causing the model to degenerate into its baseline.}

\textbf{Performance Comparison on TAG-PEDES(A).}
To evaluate the task of text-based aerial person retrieval, we construct a subset of TAG-PEDES, denoted as TAG-PEDES(A), by selecting all aerial-view images along with their corresponding textual descriptions. The evaluation results, presented in Table~\ref{tab: t2a}, show that our method continues to outperform existing methods.

\textbf{Performance Comparison on ICFG-PEDES, and RSTPReid.} 
We further evaluate TAG-CLIP on two widely used benchmarks for traditional T-PR: ICFG-PEDES, and RSTPReid. The results, shown in Table~\ref{tab: icfg} and Table~\ref{tab: rstp}, indicate that our method underperforms compared to state-of-the-art methods. As discussed in the main text, this is primarily due to: 1) in single-viewpoint settings, our proposed HR-MoE module and viewpoint decoupling strategy cannot fully demonstrate their advantages; and 2) the compared T-PR methods are specifically designed and optimized for traditional T-PR, often employing stronger baseline models or additional pretraining data, which leads to better retrieval performance. In contrast, TAG-CLIP is specifically tailored for person retrieval across hybrid image viewpoints, which presents a more complex and challenging setting by nature.
\begin{figure}[h]
  \centering
  \includegraphics[width=\linewidth]{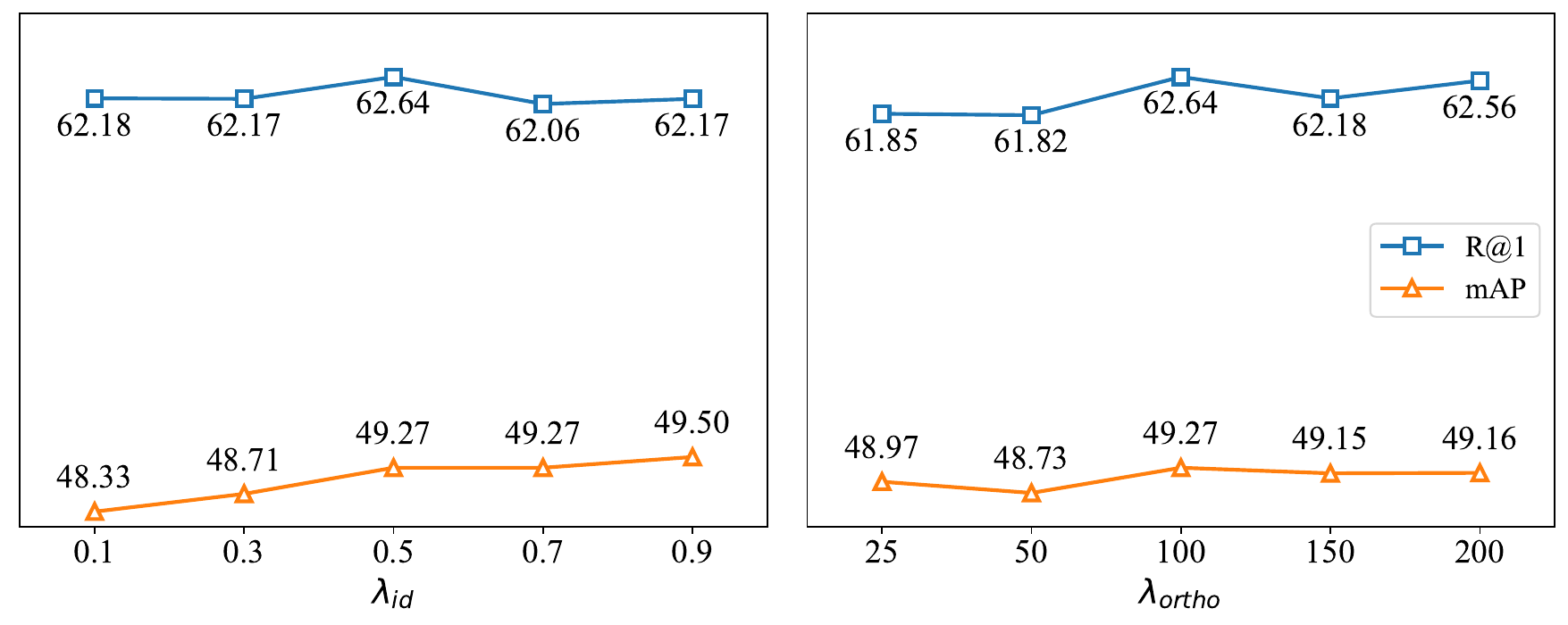}
  \caption{Hyperparameter analysis of \(\lambda_{id}\) and \(\lambda_{ortho}\).}
  \label{fig: loss weight}
\end{figure}

\begin{table}[tbp]
    \centering

    \fontsize{9}{11}\selectfont
    \setlength{\tabcolsep}{1mm}
    
    \begin{tabular}{l|cccc}
    \hline
    Methods        & R@1    & R@5    & R@10   & mAP    \\ 
    \hline\hline
    \multicolumn{5}{l}{\textit{w/o CLIP:}} \\
    \hline
    APTM~\cite{yang2023towards}        & 58.06 & 75.92 & 82.26 & 54.34 \\
    RaSa~\cite{bai2023rasa}            & 58.01 & 76.35 & 82.17 & 53.59 \\
    AUL~\cite{li2024adaptive}          & 55.24 & 74.19 & 81.20 & 52.29  \\
    \hline
    \multicolumn{5}{l}{\textit{w/ CLIP:}} \\
    \hline
    CFine~\cite{yan2023clip}           & 52.38 & 72.13 & 79.58 & -  \\
    IRRA~\cite{jiang2023cross}         & 55.85 & 74.78 & 81.80 & 52.72  \\
    TBPS-CLIP~\cite{cao2024empirical}  & 56.37 & 74.03 & 80.91 & 51.74 \\
    RDE~\cite{qin2024noisy}            & 58.57 & 75.49 & 81.95 & 54.81 \\
    AEA-FRIM~\cite{wang2025aea}        & 47.38 & 68.03 & 76.03 & 42.78 \\
    \hline
    TAG-CLIP                          & \textbf{59.60} & \textbf{77.35} & \textbf{83.40} & \textbf{55.12} \\ 
    \hline
    \end{tabular}
    \caption{Comparison with state-of-the-art methods on TAG-PEDES(A).}
    \label{tab: t2a}
\end{table}

\begin{table}[t]
    \centering
    \fontsize{9}{11}\selectfont
    \setlength{\tabcolsep}{1mm}
    \begin{tabular}{l|cccc}
    \hline
    Methods         & R@1   & R@5   & R@10   & mAP    \\ 
    \hline\hline
    \multicolumn{5}{l}{\textit{w/o CLIP:}} \\
    \hline
    LGUR~\cite{shao2022learning}               & 57.42 & 74.97 & 81.45 &   -     \\ 
    RaSa~\cite{bai2023rasa}                    & 65.28 & 80.40 & 85.12 & 41.29   \\ 
    APTM~\cite{yang2023towards}                & 68.51 & 82.99 & 87.56 & 41.22   \\ 
    AUL~\cite{li2024adaptive}                  & \textbf{69.16} & \textbf{83.32} & \textbf{88.37} &    -   \\ 
    \hline
    \multicolumn{5}{l}{\textit{w/ CLIP:}} \\
    \hline
    CFine~\cite{yan2023clip}               & 60.83 & 76.55 & 82.42 &   -   \\
    IRRA~\cite{jiang2023cross}             & 63.46 & 80.25 & 85.82 & 38.06  \\
    Propot~\cite{yan2024prototypical}      & 65.12 & 81.57 & 86.97 & 42.93  \\ 
    TBPS-CLIP~\cite{cao2024empirical}      & 64.52 & 80.03 & 85.39 & 39.54  \\ 
    RDE~\cite{qin2024noisy}                & 67.68 & 82.47 & 87.36 & 40.06  \\ 
    \hline
    TAG-CLIP                               & 66.78 & 81.36 & 86.23 & 43.60 \\
    \hline
    \end{tabular}
    \caption{Comparison with state-of-the-art methods on ICFG-PEDES.}
    \label{tab: icfg}
\end{table}

\begin{table}[t]
    \fontsize{9}{11}\selectfont
    \setlength{\tabcolsep}{1mm}
    \centering

    \begin{tabular}{l|cccc}
    \hline

    Methods           & R@1   & R@5   & R@10  & mAP   \\ 
    \hline\hline
    \multicolumn{5}{l}{\textit{w/o CLIP:}} \\
    \hline
    RaSa~\cite{bai2023rasa}                     & 66.90 & 86.50 & 91.35 & 52.31 \\ 
    APTM~\cite{yang2023towards}                 & 67.50 & 85.70 & 91.45 & \textbf{52.56} \\ 
    AUL~\cite{li2024adaptive}                   & \textbf{71.65} & \textbf{87.55} & \textbf{92.05} &  -   \\ 
    \hline
    \multicolumn{5}{l}{\textit{w/ CLIP:}} \\
    \hline
    CFine~\cite{yan2023clip}                    & 50.55 & 72.50 & 81.60 & -  \\
    IRRA~\cite{jiang2023cross}                  & 60.20 & 81.30 & 88.20 & 47.17  \\
    Propot~\cite{yan2024prototypical}           & 61.87 & 83.63 & 89.70 & 47.82  \\ 
    TBPS-CLIP~\cite{cao2024empirical}           & 62.10 & 81.90 & 87.75 & 48.00  \\ 
    RDE~\cite{qin2024noisy}                     & 65.35 & 83.95 & 89.90 & 50.88 \\ 
    \hline
    TAG-CLIP                                    & 63.85 & 82.50 & 88.95 & 50.22 \\
    \hline
    \end{tabular}
    \caption{Comparison with state-of-the-art methods on RSTPReid.}
    \label{tab: rstp}
\end{table}

\begin{table}[tbh]
    \centering
    \begin{tabular}{ccc}
    \hline

    Model & Total Params & Inference Speed \\ 
    \hline\hline
    Baseline         & 155M &  412 Samples/s\\
    TAG-CLIP         & 226M (+71M) & 242 Samples/s (×0.59) \\

    \hline
    \end{tabular}
    \caption{Comparison of resource consumption and inference speed between the baseline model and TAG-CLIP.}
    \label{tab: resource consumption}
    
\end{table}

\subsection{Ablation on HR-MoE}
We investigate the impact of HR-MoE block placement within the image encoder, with results summarized in Table~\ref{tab: ablation on MoE layers}. The results reveal two key findings.
(1) An insufficient number of HR-MoE blocks (\emph{e.g.,} one) or an excessive number (\emph{e.g.,} five) hinders the module's ability to function effectively, resulting in suboptimal performance.
(2) When a moderate number of HR-MoE blocks is employed, inserting them into the middle layers outperforms placements in either the early or late layers.
This phenomenon can be attributed to the hierarchical processing nature of ViT. Specifically, early layers primarily learn low-level visual cues, where view-specific information is not yet fully encoded, making view decoupling less effective. In contrast, the later layers capture high-level semantic features, where view-specific biases are already entangled, limiting the benefit of further decoupling. 

Additionally, we study the influence of the number of experts \(N_e\) and the number of selected experts \(K\), with the results presented in Table~\ref{tab: ablation on expert}. The model achieves optimal performance when \(N_e = 6\) and \(K = 5\), a setting in which all experts within each expert group are selected and the feature-level router's role shifts to assigning appropriate weights to each expert within the group.

\begin{table}[tb]
    \centering
    \fontsize{9}{11}\selectfont
    \setlength{\tabcolsep}{1mm}

    \begin{tabular}{c|c|cccc}
    \hline
    \#HR-MoE & Location & R@1    & R@5    & R@10   & mAP    \\ 
    \hline\hline
    $1$            & $1$      & 61.58 & 78.50 & 84.41 & 48.94 \\
    $1$            & $7$      & 61.75 & 78.88 & 84.36 & 48.87 \\
    $1$            & $12$     & 61.60 & 78.37 & 84.16 & 48.38  \\
    \hline
    $3$            & $1$--$3$    & 61.45 & 78.71 & 84.29 & 49.03 \\
    $3$            & $7$--$9$    & \textbf{62.64} & \textbf{79.10} & \textbf{84.81} & \textbf{49.27} \\
    $3$            & $10$--$12$  & 61.31 & 78.42 & 84.16 & 48.36 \\
    \hline
    $5$            & $1$--$5$    & 61.49 & 78.57 & 84.29 & 48.96 \\
    $5$            & $7$--$12$   & 61.55 & 78.31 & 84.09 & 48.63 \\

    \hline
    \end{tabular}
    \caption{Ablation study on the number and location of HR-MoE blocks.}
    \label{tab: ablation on MoE layers}
\end{table}

\begin{table}[tb]
    \centering
    \fontsize{9}{11}\selectfont
    \setlength{\tabcolsep}{1mm}
    \begin{tabular}{c|c|cccc}
    \hline
    \#Experts & \#Selected Experts & R@1    & R@5    & R@10   & mAP    \\ 
    \hline\hline
    $4$            & $3$        & 61.91 & 78.90 & 84.58 & 48.81 \\
    $5$            & $4$        & 62.49 & 79.11 & 84.61 & 49.07 \\
    \hline
    $6$            & $4$        & 62.31 & \textbf{79.15} & 84.76 & 49.07 \\
    $6$            & $5$        & \textbf{62.64} & 79.10 & \textbf{84.81} & \textbf{49.27} \\
    \hline
    $7$            & $6$        & 62.11 & 79.11 & 84.71 & 48.89 \\

    \hline
    \end{tabular}
    \caption{Ablation study on the number of Experts and Selected Experts in HR-MoE.}
    \label{tab: ablation on expert}
\end{table}

\subsection{Hyperparameter Analysis}
The hyperparameters \(\lambda_{id}\) in Eq.~15 and \(\lambda_{ortho}\) in Eq.~16 are introduced to balance different losses. We vary their values and report the results in Figure~\ref{fig: loss weight}. Overall, the results demonstrate the robustness of our method. We set \(\lambda_{id} = 0.5\) and \(\lambda_{ortho} = 100\) empirically.

\subsection{Resource Consumption Analysis} 
Compared to the baseline, TAG-CLIP introduces an additional $71$M parameters, reaching a total of $226$M as present in Table~\ref{tab: resource consumption}. However, thanks to the sparse activation mechanism in HR-MoE, only a portion of them are activated during inference. Although the inference speed drops to $0.59$× that of the baseline, mainly due to the involvement of more visual expert. However, TAG-CLIP achieves significantly improved retrieval performance, with gains of $3.55\%$ in R@1 and $1.22\%$ in mAP shown in the Tabel~4 of the main text.

\end{document}